\documentclass{article}

\usepackage{microtype}
\usepackage{graphicx}
\usepackage{subcaption}
\usepackage{booktabs} 

\usepackage{hyperref}

\usepackage[accepted]{icml2026}

\usepackage{amsmath}
\usepackage{amssymb}
\usepackage{mathtools}
\usepackage{amsthm}

\usepackage[capitalize,noabbrev]{cleveref}

\theoremstyle{plain}

\theoremstyle{definition}

\theoremstyle{remark}

\usepackage[textsize=tiny]{todonotes}

\usepackage{amsmath,amsfonts,bm}

\newcommand{\lh}{\hat{\lambda}}

\def\eqref#1{equation~\ref{#1}}

\def\1{\bm{1}}

\newcommand{\Y}{\mathcal{Y}}

\newcommand{\R}{\mathcal{R}}

\DeclareMathAlphabet{\mathsfit}{\encodingdefault}{\sfdefault}{m}{sl}
\SetMathAlphabet{\mathsfit}{bold}{\encodingdefault}{\sfdefault}{bx}{n}

\def\gD{{\mathcal{D}}}

\def\sP{{\mathbb{P}}}

\newcommand{\E}{\mathbb{E}}

\usepackage{enumitem}
\usepackage{xcolor}

\Crefname{equation}{Eq.}{Eqns.}
\Crefname{figure}{Fig.}{Figs.}
\Crefname{table}{Tab.}{Tabs.}
\Crefname{section}{§}{Sections}
\Crefname{appendix}{§}{Appendix}
\Crefname{algorithm}{Algo.}{Algorithms}

\definecolor{TabRed}{HTML}{D62728}     
\definecolor{TabGreen}{HTML}{2CA02C}   
\definecolor{TabBlue}{HTML}{1F77B4}    
\definecolor{TabOrange}{HTML}{FF7F0E}  
\definecolor{TabGrey}{HTML}{7F7F7F}    

\newcommand{\tabgreenline}{{\tikz[baseline=-0.5ex] \draw[very thick, color=TabGreen] (0., 0.) -- (0.3, 0.);}}
\newcommand{\tabblueline}{{\tikz[baseline=-0.5ex] \draw[very thick, color=TabBlue] (0., 0.) -- (0.3, 0.);}}

\newcommand{\tabgreydashedline}{{\tikz[baseline=-0.5ex] \draw[very thick, dashed, color=TabGrey] (0., 0.) -- (0.3, 0.);}}

\newcommand{\Is}{\mathcal{I}_{\text{s}}}
\newcommand{\Ius}{\mathcal{I}_{\text{us}}}
\newcommand{\Iflag}{\mathcal{I}_{\Phi}}

\icmltitlerunning{Online Safety Monitoring for LLMs}
\begin{document}

\twocolumn[
  \icmltitle{Online Safety Monitoring for LLMs}



  \icmlsetsymbol{equal}{*}

  \begin{icmlauthorlist}
    \icmlauthor{Mona Schirmer}{amlab}
    \icmlauthor{Metod Jazbec}{amlab}
    \icmlauthor{Alexander Timans}{amlab}
    \icmlauthor{Christian Naesseth}{amlab}
    \icmlauthor{Maja Waldron}{wis}
    \icmlauthor{Eric Nalisnick}{jh}
  \end{icmlauthorlist}

  \icmlaffiliation{amlab}{UvA Bosch-Delta Lab, University of Amsterdam}
  \icmlaffiliation{wis}{University of Wisconsin Madison}
  \icmlaffiliation{jh}{Johns Hopkins University}

  \icmlcorrespondingauthor{Mona Schirmer}{m.c.schirmer@uva.nl}

  \icmlkeywords{Machine Learning, ICML}

  \vskip 0.3in
]



\printAffiliationsAndNotice{}  

\begin{abstract}
Despite alignment training, LLMs remain prone to generating unsafe outputs at deployment time. Monitoring outputs online and raising an alarm when safety can no longer be assumed is therefore critical.
We study a simple real-time monitor that turns a verifier signal from an external model into an alarm decision by thresholding, with the threshold calibrated via risk control. In experiments on mathematical reasoning and red teaming datasets, we show that this simple design is competitive with more advanced monitors based on sequential hypothesis testing.
\end{abstract}

\section{Introduction}
\label{sec:intro}

Large Language Models (LLMs) have become integrated into our everyday lives as search engines \citep{jin2025search,xiong2024search}, coding assistants \citep{zhao2023survey}, and companions \citep{zhang2025rise}. As their applicability grows, so does the potential harm caused by malicious LLM outputs. Despite remarkable performance across a wide range of tasks, LLMs remain prone to generating hallucinated, factually incorrect \citep{ravichander2025halogen}, or harmful output \citep{yu2025survey} when deployed.

LLM safety research has addressed these challenges using better alignment strategies during training \citep{bai2022training}, and rigorous offline evaluation before deploying \citep{wang2023decodingtrust}. Nevertheless, these pre-deployment safety measures cannot account for all possible prompt scenarios, and the risk of harmful outputs remains. There has thus been substantial effort into developing guardrails at inference time \citep{inan2023llama,sharma2025constitutional,baker2025monitoring} -- though many of them are designed for post-hoc detection. 
We argue that such monitors should be able to operate in an online stream setting, detecting unsafe content as it is produced, and stopping harmful generations in real-time
\citep{wang2025pro2guard,li2025judgment}.

The inherent challenges of the online monitoring setting are the unavailability of safety labels, and the fast reaction time required for effective detection. In an ideal setting, every output is reviewed by a group of human experts who judge whether it is harmful, incorrect, or otherwise problematic. In the real world, however, we only get an \emph{approximate} signal that informs us of the safety of the current output, such as the predictive probability from a safeguard classifier. The task of an online monitor is then to translate this signal into a binary decision---raise an alarm or remain silent---as the LLM produces its output. The monitor itself comes with two main risks: \emph{(i)} raising false alarms unnecessarily interrupts user experience and functionality of the LLM, and \emph{(ii)} failing to detect true alarms gives a sense of false security and renders the monitor useless. 

In this paper, we study the online monitoring setting covering several safety risks \citep{weidinger2023sociotechnical}: factual correctness, toxicity and malicious use. We deploy a simple statistical framework based on \emph{risk control} \citep{angelopoulos2022conformal} that converts any safety signal into a binary decision rule, and offers statistical guarantees on the false alarm  or missed detection rate. The framework is universally applicable to different monitoring purposes and can leverage arbitrary proxy signals. Through experiments on mathematical problem solving and red teaming conversations,  
we show that our simple approach is competitive with more involved methods \citep{sadhuka2025valuator}, while detecting failures earlier in the generation process.

\section{Problem Setting}
We consider the problem of monitoring the safety of an LLM's output as it unfolds. Let $t = 0, 1, \dots$ be a time index. At $t = 0$, the LLM is given a user prompt $x \sim P_x$ sampled from a prompt distribution. The LLM then produces an output sequence $o_{1:T} = (o_1, \dots, o_T)$ of variable length $T$, where each $o_t$ is sampled autoregressively from the model's generative distribution ${P_\theta(o_t \mid x, o_{1:t-1})}$. Depending on the use case, each $o_t$ may represent a token of the output text, a step in the reasoning chain, or a response of the LLM to the user. Let $y \in \Y = \{0, 1\}$ be the safety variable, sampled from $P(y \mid o_{1:T})$, which captures the safety of the output sequence, with $y = 1$ indicating a safe state (e.g. correct, harmless output) and $y = 0$ indicating an unsafe state (e.g. incorrect, harmful output). The goal of an online monitor is to identify an unsafe sequence as early as possible, while continuously inspecting $o_{1:t}$ as it unfolds.

\section{Online Monitoring via Risk Control}
We study a simple statistical baseline for monitoring LLM safety at inference time. More formally, given a stream of safety signals $s_{1:t}$, we seek a decision rule -- i.e., a well-calibrated threshold $\lambda$ -- that maps $s_{1:t}$ to a binary judgment of whether the output is safe. We discuss appropriate choices of safety signals in \Cref{sec:signal} and state the monitor function in \Cref{sec:monitor}. In \Cref{sec:risk}, we define the risks of the monitor we wish to control and lastly, in \Cref{sec:control}, provide two options on how to determine a decision threshold $\lambda$ which guarantees these risks remain controlled \emph{in expectation} or \emph{with high probability}. 

\subsection{Safety Signal}
\label{sec:signal}
To assess the safety of an output, we rely on an imperfect signal $s_t$ that carries information about $y$. For example, $s_t$ may be the predictive probability of an external verifier model $p_{\psi}(y \mid x, o_{1:t})$ that assesses safety given all currently observed outputs $o_{1:t}$. In mathematical reasoning, $p_{\psi}$ may correspond to a process reward model (PRM) \citep{lightman2024let, wang2024math, prmlessons}; in content moderation, it can be an LLM safeguard \citep{inan2023llama, zeng2024shieldgemma}. While external models explicitly trained for this binary prediction are effective, they are also potentially expensive to deploy alongside the LLM generator. Internal signals from the generating LLM $p_{\theta}(o_t \mid x, o_{1:t-1})$ have therefore been explored as a cheaper alternative \citep{agarwal2026unreasonable, kossen2024semantic}. We conduct an ablation on this trade-off in \Cref{app:signal}.

\subsection{Safety Monitor}
\label{sec:monitor}
The monitor processes the signal sequence $s_{1:t}$ online and emits a binary decision at each time step $t=1, 2\dots $, with $\Phi_t = 1$ indicating the output has been flagged as unsafe. We define it as a stopping rule governed by a single threshold $\lambda$: the monitor raises an alarm at the first step $k$ at which $s_k$ falls below $\lambda$. Formally, we write
\begin{align}
\label{eq:monitor}
    \Phi_t := \mathbf{1} \{\exists k,\ 1 \leq k \leq t : s_k < \lambda\}.
\end{align}
Upon raising an alarm at time $t$, downstream interventions may be triggered, for instance, by halting generation, escalating to a stronger verifier, or invoking a human-in-the-loop review. We emphasize the simplicity of the rule in \Cref{eq:monitor}: it relies on a single, time-invariant threshold $\lambda$ applied uniformly across all signals $s_t$. How to choose $\lambda$ such that the monitor's risks are provably controlled is the subject of the remainder of this section. First, we turn to formalizing our notions of risk themselves.

\subsection{Risks of Monitors}
\label{sec:risk}

As discussed in \Cref{sec:intro}, a monitor is exposed to two complementary error modes, mirroring the type~I and type~II errors of classical hypothesis testing \citep{kaur2017type}. The \emph{false alarm risk} captures the probability of flagging a sequence that is in fact safe (type~I error, $\R^{I}$), while the \emph{missed detection risk} captures the probability of failing to raise an alarm on a sequence that is unsafe (type~II error, $\R^{II}$). We focus on the false alarm risk henceforth and refer to \Cref{app:missed-detection} for methodology and experiments on the missed detection risk.

\paragraph{False Alarm Risk.} The probability that the monitor governed by a specific $\lambda$ raises an alarm at some step $t$ given that the underlying sequence is safe is denoted
\begin{align}
    \label{eq:false_alarm_risk}
    \R^{I}(\lambda) = \sP(\exists\, t \geq 1 : s_{t} < \lambda \mid y=1).
\end{align}
Note that this is an expectation of the binary loss $\ell(s_{1:t}, \lambda)= \mathbf{1}\{\exists\, t \geq 1 : s_{t} < \lambda\}$ taken over safe samples only ($y=1$), so $\R^{I}(\lambda) \in [0,1]$. $\R^{I}(\lambda)$ is \emph{monotonically increasing} in $\lambda$: raising the threshold makes the monitor more eager to flag, increasing false alarms.

\subsection{Controlling Monitoring Risks}
\label{sec:control}
Having defined the risks of interest in \Cref{sec:risk}, we now turn to the central calibration question: how should the threshold $\lambda$ be chosen so that the resulting monitor provably controls a targeted risk $\R \in \{\R^I, \R^{II}\}$ at user-specified risk level $\epsilon \in (0, 1)$? We consider access to a labeled held-out calibration dataset $\gD_{\text{cal}} = \{(x^{(i)}, o_{1:T}^{(i)}, y^{(i)})\}_{i=1}^n$ of $n$ calibration samples, drawn exchangeably from the same distribution as the deployment data. We can then evaluate empirical risks $\hat{\R}(\lambda; \gD_{\text{cal}})$ for any candidate threshold $\lambda \in \Lambda$. Two notions of control are natural in this setting, yielding two distinct calibration procedures that differ in the strength of their statistical guarantees.

\paragraph{Control in expectation.} The first option, \emph{conformal risk control} \citep{angelopoulos2022conformal},  provides a way to find a threshold $\hat{\lambda}$ on a given calibration dataset $D_{\text{cal}}$, such that the risk, i.e. expected loss, of future test points is bounded \emph{on average} over draws of possible calibration sets:
\begin{align}
\label{eq:expectation}
    \E_{\gD_{\text{cal}}}[\R(\lh)] \leq \epsilon.
\end{align}
Exploiting that the loss $\ell$ is non-decreasing in $\lambda$ for \Cref{eq:false_alarm_risk}, the guarantee is achieved by selecting the largest threshold whose finite-sample-corrected empirical risk still lies below the target level:
{\small
\begin{align} 
\label{eq:lambda_crc}
    \lh_{\mathrm{CRC}} := \max\left\{ \lambda \in \Lambda \,:\, \frac{n}{n+1}\, \hat{\R}(\lambda; \gD_{\text{cal}}) + \frac{1}{n+1} \leq \epsilon \right\}.
\end{align}
}

\paragraph{Control with high probability.} The second option, based on \citet{bates2021distribution}, provides a stronger guarantee: for a user-specified confidence level $(1 - \delta)$ the same risk is bounded for \emph{all but a $\delta$-fraction of calibration draws}, formally
\begin{align}
\label{eq:high_prob}
    \sP_{\gD_{\text{cal}}}\!\left(\R(\lh) \leq \epsilon\right) \geq 1 - \delta.
\end{align}
The construction proceeds in two steps. First, a $(1-\delta)$ upper confidence bound (UCB) $U(\lambda, n, \delta)$  on the empirical risk $\hat{\R}(\lambda; \mathcal{D}_{\text{cal}})$ is computed on the calibration set. It follows ${\sP_{\mathcal{D}_{\text{cal}}}(\R(\lambda) \leq U(\lambda, n, \delta)) \geq 1 - \delta}$ for all $\lambda \in \Lambda$. Then, the decision threshold is taken as the largest value for which this upper bound still meets the target rate $\epsilon$, hence
\begin{align}
\label{eq:lambda_ucb}
    \lh_{\mathrm{UCB}} := \max\{\lambda \in \Lambda \,:\, U(\lambda, n, \delta) \leq \epsilon\}.
\end{align}
Because the true risk $\R(\lambda)$ lies below the bound with probability at least $(1 - \delta)$, any $\lambda$ selected by this rule inherits the same guarantee, yielding high-probability control. We use the Hoeffding-Bentkus bound \citep{bentkus2004hoeffding} as UCB. 

In contrast to in-expectation control, high-probability control offers a stronger deployment guarantee at the cost of a more conservative threshold and, typically, the need for a larger calibration set size $n$. Thus, in practice, UCB may be preferred when the cost of a single safety violation is high and sufficient calibration samples are available. We refer to the monitor instantiated by \Cref{eq:lambda_crc} as CRC and \Cref{eq:lambda_ucb} as UCB respectively.

\section{Experiments}
We evaluate the risk-controlling monitor on two safety use cases: (i) factuality by monitoring  step-level correctness for mathematical reasoning (\Cref{sec:exp_factuality}) and (ii)  malicious use and toxicity, by monitoring multi-turn conversations with users (\Cref{sec:exp_harm}). Finally, \Cref{app:signal} studies the cost–performance trade-off when using cheap internal model signals in place of external verifiers. Our code is available at \url{https://github.com/monasch/llm-monitor}.

\paragraph{Baselines} We compare the simple risk control monitor to two versions of the e-valuator \citep{sadhuka2025valuator}. All monitors observe the same signal $s_t$ and have access to the same calibration set, but differ in the monitor function. Importantly, instead of thresholding the raw signal by a single $\lambda$ (\Cref{eq:monitor}), e-valuators learn $T$ density estimators (one for each step $t$) on the signal sequence $s_{1:T}$. The resulting \emph{e-process} is used as evidence signal for a sequential hypothesis test. Details on e-valuators are in \Cref{app:e_valuator}.

\paragraph{Metrics} We assess the monitors across 3 metrics. Over
$i = 1, \dots, N$ test sequences, let $\Is = \{i : y^{(i)} = 1\}$,
$\Ius = \{i : y^{(i)} = 0\}$, and $\Iflag = \{i : \exists t, \Phi^{(i)}_t = 1\}$ denote
the safe, unsafe, and flagged sequences, respectively.
\begin{itemize}[label=\textendash, noitemsep, topsep=0pt, leftmargin=*, labelsep=4pt]
    \item \textbf{False Alarm Rate}, the share of safe sequences that are incorrectly flagged,
    $\hat{R}^{I}(\lh) = |\Iflag \cap \Is| \,/\, |\Is|$,
    i.e., the empirical estimate of the false alarm risk in \Cref{eq:false_alarm_risk}. Lower is better.
    \item \textbf{Power}, the share of unsafe sequences that are correctly flagged,
    $1 - \hat{R}^{II}(\lh) = |\Iflag \cap \Ius| \,/\, |\Ius|$,
    i.e., one minus the empirical estimate of the missed detection risk (\Cref{eq:missed_detection},\Cref{app:missed-detection}). Higher is better.
    \item \textbf{Detection Delay}, the average fraction of steps the model processes before an alarm is correctly raised,
    $\hat{DD} = \frac{1}{|\Iflag \cap \Ius|} \sum_{i \in \Iflag \cap \Ius} t_{\min}^{(i)} / T^{(i)}$,
    where $t_{\min}^{(i)} = \min\{t : s_t^{(i)} < \lh\}$ is the alarm time and $T^{(i)}$ is the sequence length. Lower is better.
\end{itemize}

\subsection{Monitoring for Factuality}
\label{sec:exp_factuality}
In our first experiment, we consider the safety risk caused by LLMs 
providing factually incorrect statements \citep{weidinger2021ethical, ji2023survey}. 
We focus on mathematical reasoning, where incorrect solutions can cause 
false belief in the user especially when they overestimate LLM competence 
\citep{weidinger2021ethical, steyvers2025large}.

\paragraph{Setup} We use the MATH dataset \citep{hendrycksmath2021} to monitor an LLM's capacity to produce correct step-by-step solutions to mathematical problems. We employ two generating LLMs of varying problem-solving capacity: Claude Haiku 4.5 \citep{anthropic2025haiku45} solves $90\%$ of problems correctly, whereas Mistral-7B-Instruct-v0.3 \citep{jiang2023mistral} solves only $26\%$. We use OpenAI's o3-mini \citep{openai2025o3mini} to compare each generator's final response with the ground-truth solution, providing the $y$ labels. As signal $s_t$, we use the step-wise output probability of Qwen2.5-Math-PRM-7B \citep{prmlessons}.
\begin{figure}[t]
    \centering
    \includegraphics[width=\columnwidth]{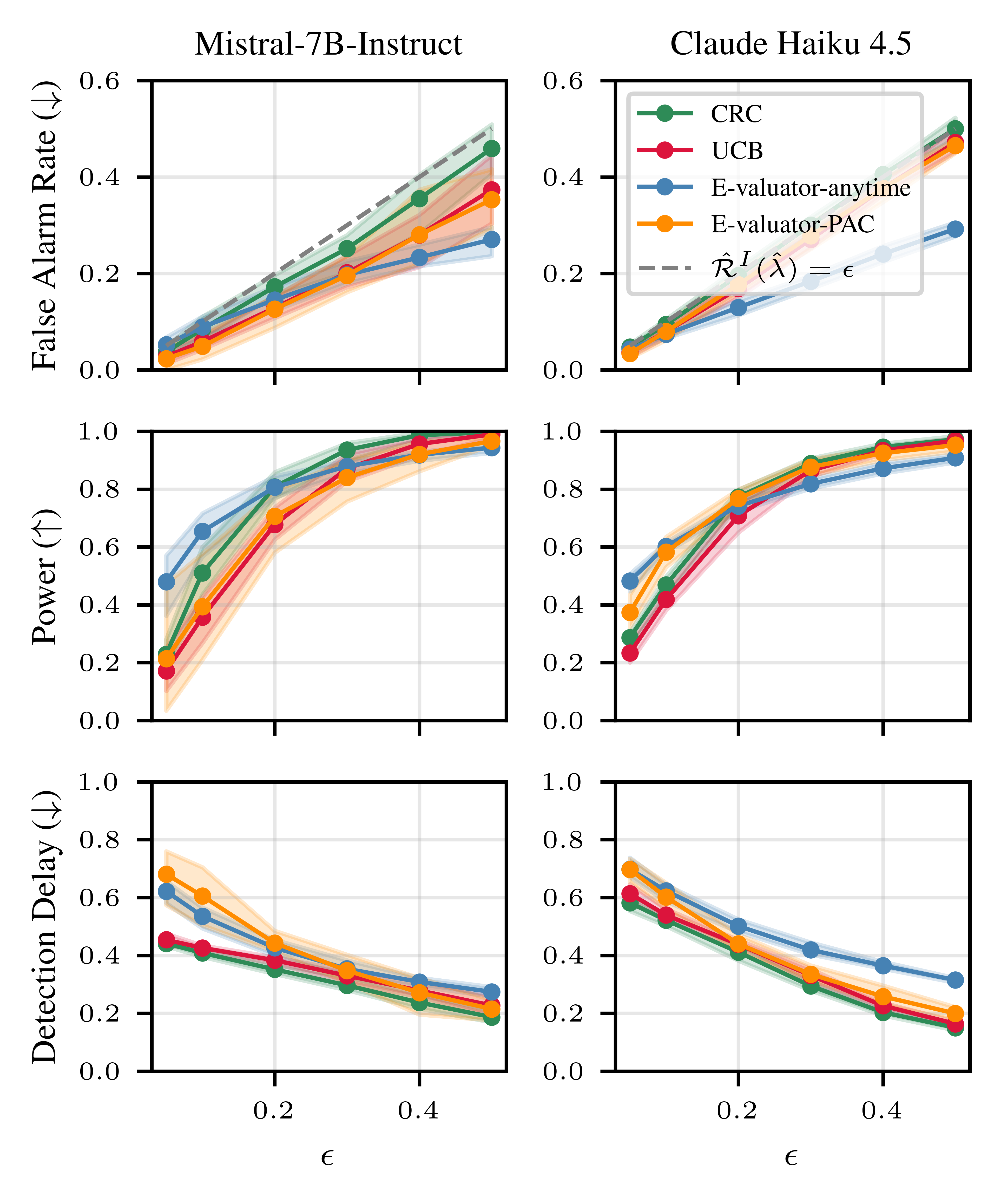}
    \vspace{-20pt}
    \caption{Monitoring factuality on mathematical reasoning (MATH): CRC and UCB calibrate a single threshold to turn a PRM score into an alarm, yet detect incorrect answers earlier (\textit{third row}) than e-valuators, which train several density estimators.}
    
    \label{fig:monitors_math}
\end{figure}

\paragraph{Risk remains empirically controlled.}
\Cref{fig:monitors_math} shows the false alarm rate, power, and detection delay when controlling the monitor's false alarm risk (\Cref{eq:false_alarm_risk}) across varying tolerance levels $\epsilon$. The monitors with high-probability risk control (\Cref{eq:high_prob}) -- UCB, e-valuator-anytime, and e-valuator-PAC -- successfully control the false alarm rate (\textit{first row}), with the $1-\delta$ confidence interval over 10 runs (\textit{shaded region}) lying below the identity line; the only exception is e-valuator-anytime (\tabblueline), which violates the bound at $\epsilon \in \{0.05, 0.1\}$ on the Mistral model. The CRC monitor likewise satisfies its in-expectation guarantee (\tabgreenline $ < $ \tabgreydashedline).

\paragraph{Risk controlling monitors raise alarms earlier.}
Despite their simplicity, CRC and UCB are surprisingly on par with the more complex e-valuator monitors. Looking at power (\textit{second row}, \textit{how many} incorrect sequences are detected), e-valuator-anytime has the highest at small tolerance levels, though this comes at the cost of risk violation. E-valuator-PAC and UCB are on par. As expected, the less conservative CRC yields higher power than UCB. Turning to detection delay (\textit{third row}, \textit{how quickly} the detected sequences are flagged) yields an interesting picture: although the e-valuator monitors detect more incorrect sequences, they do so much later. CRC and UCB instead flag incorrect sequences after about half the sequence, with CRC slightly ahead of UCB. Earlier detection is desirable in practice, as it limits the user's exposure to harmful outputs and reduces the token-generation cost.

\subsection{Monitoring for Harmlessness}
\label{sec:exp_harm}
In our next experiment, unsafety stems not from incorrect outputs, but from the LLM complying with malicious requests or producing discriminatory and toxic content.

\begin{figure}[t]
    \centering
    \includegraphics[width=\columnwidth]{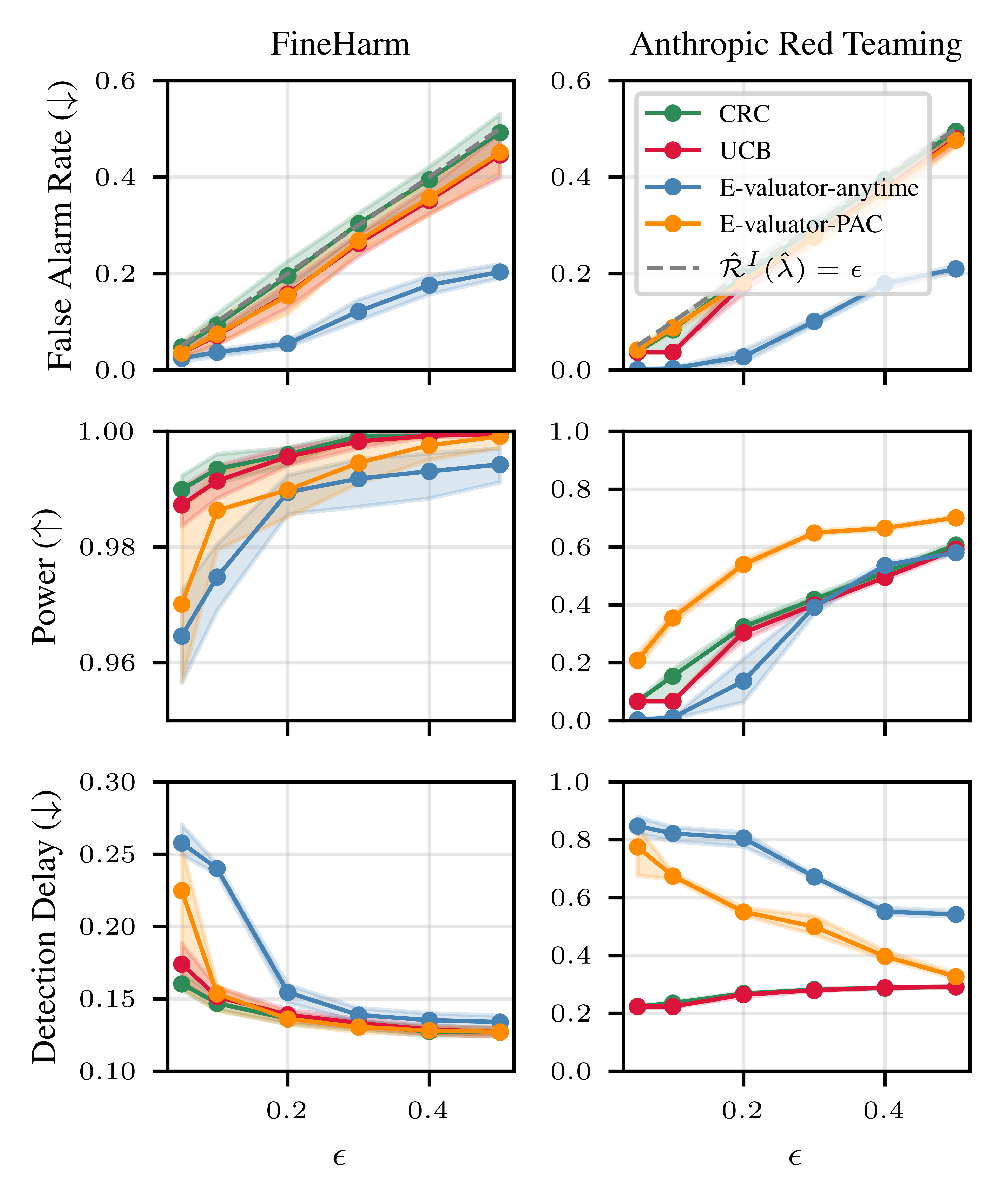}
    \vspace{-20pt}
    \caption{Monitoring harmlessness: All monitors maintain false alarm control (\textit{first row}); power (\textit{second row}) varies strongly based on whether the safeguard verifier is trained for the specific online detection (FineHarm, \textit{first column}) or not (Red Teaming, \textit{second column}). Despite its simplicity, risk controlling monitors (CRC, UCB) detect harmful output earlier (\textit{third row}).}
    
    \label{fig:monitors_harm}
\end{figure}

\paragraph{Setup} We evaluate our monitor on two datasets of harmful LLM interactions. The Anthropic Red Teaming data \citep{ganguli2022red} consists of multi-turn conversations between red-team members and an LLM, each rated for attack success, which provides the sequence-level safety label $y$. At each turn $t$, we feed the conversation up to the current time point, $o_{1:t}$, to Llama Guard \citep{inan2023llama} and use its predicted probability of the \texttt{safe} token as the verifier signal $s_t$. The FineHarm data \citep{li2025judgment} is built on WildGuard \citep{han2024wildguard} and WildJailbreak \citep{wildteaming2024} and comprises single-turn interactions. Here we use their SCM verifier, Qwen2.5-1.5B fine-tuned for token-level harmfulness detection, and take its predicted probability of safety given the tokens up to $t$ as signal $s_t$.

\paragraph{Better signals, better monitors.}
\Cref{fig:monitors_harm} displays results for FineHarm (first column) and Anthropic Red Teaming (second column). We highlight four findings. First, $\hat{R}^{I}$ remains controlled across monitors and datasets. Second, e-valuator-PAC achieves the highest power on Red Teaming, while CRC and UCB do so on FineHarm. Third, detection delay is again lower for the risk-controlling threshold methods. Fourth, comparing the two datasets highlights an important point: the monitor can only be as good as its verifier signal. On FineHarm, the SCM verifier is fine-tuned on the training set for token-level harmfulness, yielding strong separability (power close to 1), whereas Llama Guard's power on Red Teaming stagnates below 0.8. This underscores the need for accurate signals tailored to the use case.

\begin{figure*}[h!]
    \centering
    \includegraphics[width=\linewidth]{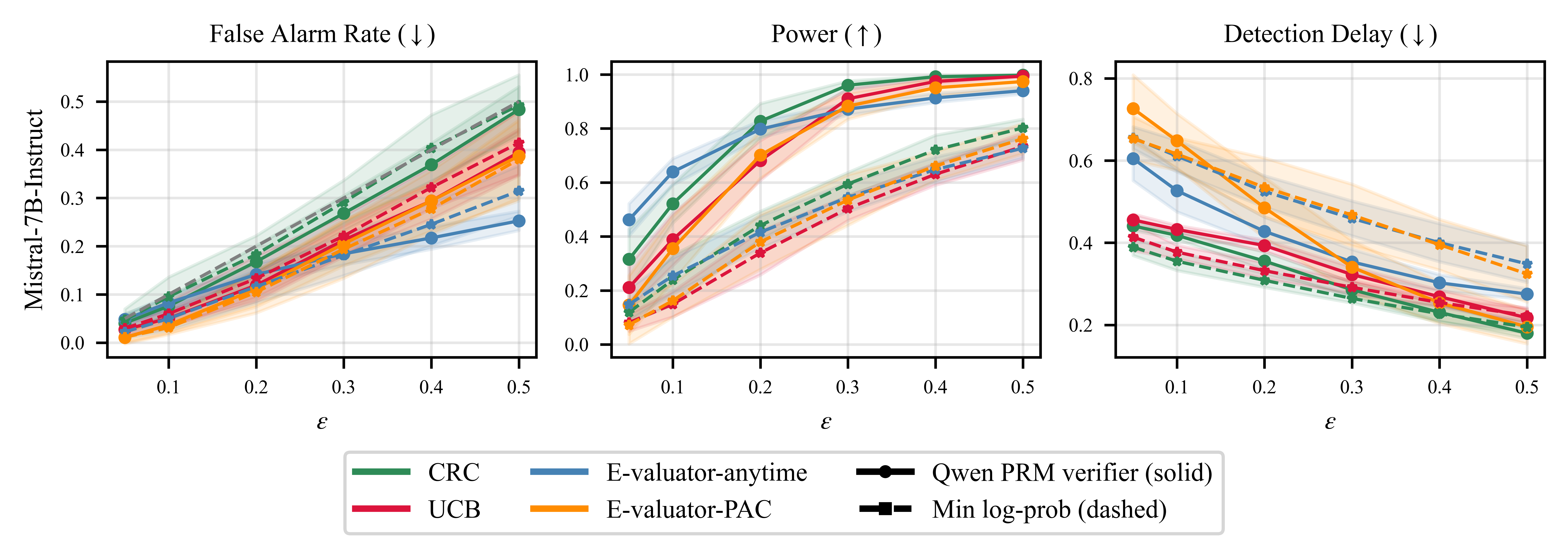}
    \vspace{-20pt}
    \caption{Signal ablation of token log-probabilities vs.\ external PRM: False alarm rate, power, and detection delay on Mistral-7B-Instruct (MATH) across target levels $\varepsilon$; solid curves use the Qwen PRM, dashed curves use the per-step minimum token log-probability. The free log-prob signal yields substantially lower power across all four monitors.}
    \label{fig:logprobs}
\end{figure*}
\subsection{Trading Monitor Efficiency for Performance}
\label{app:signal}

Running an external verifier alongside the generator may be too expensive in a practical deployment setting: it adds a second forward pass at every step. A natural question is therefore whether a much cheaper signal — one the generator already produces for free — can substitute for it, and we examine this in our next experiment.

\paragraph{Set-up.} We run a signal ablation on Mistral-7B-Instruct reasoning chains on the MATH dataset. In place of the Qwen PRM score, we use the generator's own token log-probabilities\footnote{Besides token log-probabilities, we also considered measures that capture the full token distribution at every step, e.g., self-certainty \cite{kang2026scalable}, but found them to perform on-par with the (simpler) token log-probability signal in our experiments.}  as the safety signal. Unlike the external verifier, this signal incurs practically no additional computational cost at inference time. Concretely, we found the per-step minimum over token log-probabilities to be the most informative aggregation. Formally, let $w_j$ denote the $j$-th output token of the generated chain, so that $\log p_\theta(w_j \mid w_{1:j-1}, x)$ is the generator's log-probability of that token given all preceding tokens and the prompt $x$. For a step $k$ spanning tokens $[j^{(k)}_\text{start}, j^{(k)}_\text{end}]$, we set $s_k = \min_{j \in [j^{(k)}_\text{start},\, j^{(k)}_\text{end}]} \log p_\theta(w_j \mid w_{1:j-1}, x)$, so that steps containing an unusually uncertain token receive a more negative (more unsafe) score. We feed this signal into the same set of monitors and sweep the target false-alarm level $\varepsilon \in \{0.05, 0.1, 0.2, 0.3, 0.4, 0.5\}$.

\paragraph{Token log-probabilities are a weaker signal than an external PRM.} \Cref{fig:logprobs} reports false alarm rate, power, and detection delay for the log-prob signal (dashed curves) against the PRM-based runs (solid curves). The middle column makes the trade-off between verifier cost and signal strength explicit: at matched false-alarm rates, the PRM-based monitors achieve substantially higher power than their log-prob counterparts — for example, near $\varepsilon = 0.3$ the PRM variants already exceed 0.9 power while the log-prob counterparts sit around 0.5. The generator's own token-level confidence is therefore a meaningfully weaker correctness signal than a dedicated PRM, and the gap is large enough that the cost savings come at a real monitoring-quality cost.

\section{Conclusion, Limitations and Future Work}
We argue for monitoring LLMs in real-time, enabling intervention as soon as safety can no longer be ensured. We compared statistical frameworks providing guarantees of differing strength, and found that calibrating a single threshold on a proxy signal is a simple yet effective approach to obtaining monitoring guarantees during deployment.

\paragraph{Limitations and Future Work.}
Calibrating a single time-invariant threshold on a verifier signal is attractive for deployment: it adds negligible computational overhead (e.g. no additional density estimator required) and imposes light restrictions on the calibration data (e.g. no sufficient coverage of varying length sequence necessary). However, it has two key limitations. First, the monitor is only as good as its signal — inheriting the verifier's limitations in terms of informativeness, deployment cost, and adversarial robustness. Future work could address these limitations by (i) combining multiple signals into a more informative and robust statistic; (ii) identifying the best accuracy-cost trade-off for a given safety risk \citep{gui2024conformal,kaddour2026agentic}; or (iii) issuing additional targeted safety checks once an alarm is triggered. Second, it ignores temporal structure in the signal, since the verifier score at step $t$ may systematically depend on $t$. Future work may (iv) calibrate a per-step threshold using more advanced procedures such as Pareto testing \citep{laufer2022efficiently}.

\section*{Impact Statement}
This paper presents work whose goal is to advance the field of Machine
Learning. There are many potential societal consequences of our work, none which we feel must be specifically highlighted here.

\section*{Acknowledgments}
We would like to thank Shuvom Sadhuka and Drew Prinster for a helpful exchange on their e-valuator work. This project was generously supported by the Bosch Center for Artificial Intelligence. Eric Nalisnick did not utilize resources from Johns Hopkins University for this project.

\bibliography{bib}
\bibliographystyle{icml2026}

\newpage
\appendix
\onecolumn

\section{Related Work}

\paragraph{LLM Monitoring}
Various efforts \citep{weidinger2021ethical, weidinger2023sociotechnical, bommasani2021opportunities} have stressed the importance of monitoring LLMs at deployment time, for factuality \citep{manakul2023selfcheckgpt,chuang2024lookback}, content moderation \citep{inan2023llama, zeng2024shieldgemma} or against obfuscation \citep{korbak2025chain, baker2025monitoring, greenblatt2023ai}. For content moderation, for instance, monitoring tools are typically taking the form of safeguard classifiers that predict the harmfulness of a user prompt or an LLM output \citep{inan2023llama, sharma2025constitutional, mazeika2024harmbench, markov2023holistic}. Recent work \citep{li2025judgment} proposes online monitors that are designed for unfolding output, the setting we consider here. Such oversight models can provide strong signals that can be leveraged within a statistical framework, as we discuss in this work.

\paragraph{Statistical Monitoring Frameworks} Detecting when a model degrades at inference time or exceeds a pre-defined risk level has traditionally been studied in the context of distribution shifts. Some monitoring approaches rely on confidence sequences \citep{howard2021time, shin2023detectors} to test whether the risk remains acceptable at all deployment time steps. This has been evaluated both with \citep{podkopaev2021tracking} and without \citep{amoukou2024sequential} labels, as well as under model adaptation \citep{schirmer2025monitoring, bar2024protected}. Relatedly, e-processes \citep{ramdas2023game, ramdas2025hypothesis} have been used to track evidence of risk violations over time \citep{timans2025continuous, prinster2025watch}. In the context of LLMs, conformal prediction has been used to provide statistical trustworthiness, albeit in a classic offline setting \citep{cherian2024large, mohri2024language, quach2023conformal, gui2024conformal}. 
A related line of work on optimising reasoning budget develops dynamic statistical frameworks that yield optimal stopping rules for exiting thinking mode; however, their primary goal is efficiency \citep{wang2026conformal, you2025probabilistic, zeng2025pac, wu2025thought, jazbec2024fast}. Another line of work \citep{davidov2025calibrated,feldman2026many} uses conformal survival analysis to construct PAC-type bounds on the time-to-unsafe-sampling of a given prompt. Most closely related to our work are agentic oversight methods \citep{wang2025pro2guard}, particularly \citet{sadhuka2025valuator}, which tackles the same online monitoring setting.

\section{Missed Detection Risk}
\label{app:missed-detection}
We complement the false alarm risk of \Cref{sec:risk} with its counterpart, the missed detection risk. The two risks trade off against each other: lowering one comes at the cost of raising the other. Importantly, risk control is particularly well suited to navigating this trade-off, as it lets the practitioner choose which risk to control and at what level. This stands in contrast to sequential hypothesis tests such as e-valuators \citep{sadhuka2025valuator}, where one can only control the false alarm rate without any guarantee on power. We first formalize the missed detection risk and describe how the calibration procedures of \Cref{sec:control}
extend to it, and then present experiments on mathematical reasoning.
\paragraph{Missed Detection Risk.} The probability that the monitor never raises an
alarm given that the underlying sequence is unsafe is denoted
\begin{align}
\label{eq:missed_detection}
    \R^{II}(\lambda) = \sP(\forall\, t \geq 1 : s_{t} \geq \lambda \mid y=0).
\end{align}
This is an expectation of the binary loss
$\ell(s_{1:t}, \lambda) = \mathbf{1}\{\forall\, t \geq 1 : s_t \geq \lambda\}$ taken over
unsafe samples only ($y=0$), so $\R^{II}(\lambda) \in [0,1]$; equivalently,
$\R^{II}(\lambda) = 1 - \mathrm{power}(\lambda)$. In contrast to $\R^{I}$,
$\R^{II}(\lambda)$ is \emph{monotonically decreasing} in $\lambda$: raising the
threshold makes the monitor in \Cref{eq:monitor} more eager to flag, reducing missed detections at the cost
of additional false alarms.

\paragraph{Controlling the Missed Detection Risk.}
The two calibration procedures of \Cref{sec:control} extend to $\R^{II}$ with a single
modification. Because the empirical risk is now monotonically \emph{decreasing} in
$\lambda$, the target constraint is satisfied for all sufficiently large thresholds, and
we therefore select the \emph{smallest} valid $\lambda$ rather than the largest. This
yields the most permissive monitor that still meets the desired guarantee. The
empirical risk $\hat{\R}^{II}(\lambda; \gD_{\text{cal}})$ is evaluated on the unsafe
subset of the calibration data, $\{(x^{(i)}, o^{(i)}_{1:T}, y^{(i)}) \in \gD_{\text{cal}} : y^{(i)} = 0\}$,
of effective size $n_0$.
For control in expectation, the threshold is
\begin{align}
    \lh_{\mathrm{CRC}} := \min\left\{ \lambda \in \Lambda \,:\, \frac{n_0}{n_0+1}\,
        \hat{\R}^{II}(\lambda; \gD_{\text{cal}}) + \frac{1}{n_0+1} \leq \epsilon \right\},
\end{align}
which guarantees $\E_{\gD_{\text{cal}}}[\R^{II}(\lh_{\mathrm{CRC}})] \leq \epsilon$.
For control with high probability, let $U(\lambda, n_0, \delta)$ be a $(1-\delta)$ upper
confidence bound on $\hat{\R}^{II}(\lambda; \gD_{\text{cal}})$. The threshold is then
\begin{align}
    \lh_{\mathrm{UCB}} := \min\{\lambda \in \Lambda \,:\, U(\lambda, n_0, \delta) \leq \epsilon\},
\end{align}
which guarantees
$\sP_{\gD_{\text{cal}}}(\R^{II}(\lh_{\mathrm{UCB}}) \leq \epsilon) \geq 1 - \delta$.

\begin{figure*}
    \centering
    \includegraphics[width=\linewidth]{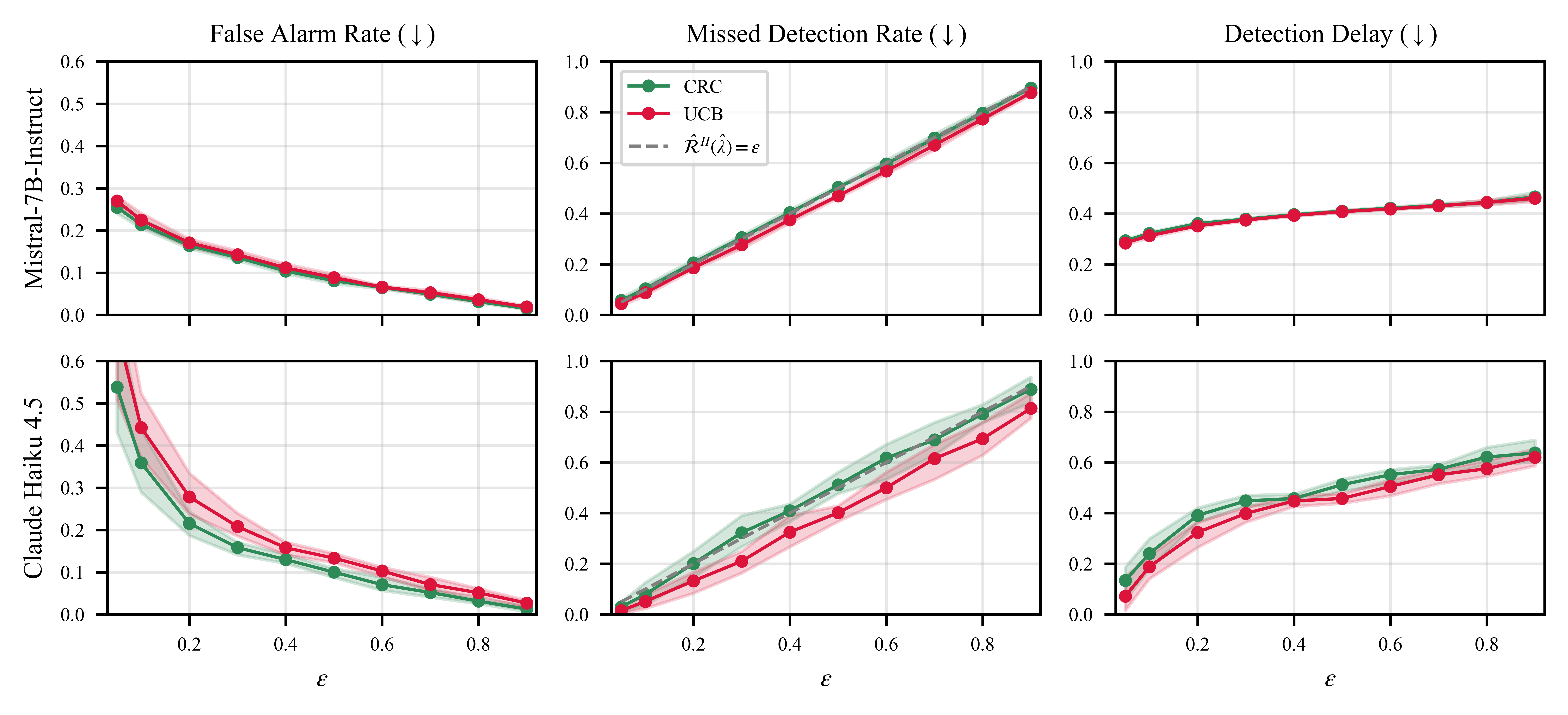}
    \vspace{-20pt}
    \caption{Monitoring performance when controlling missed detection risk $\R^{II}$ (\Cref{eq:missed_detection}) instead of false alarm risk $\R^{I}$ (\Cref{eq:false_alarm_risk}): CRC and UCB control the risk $\R^{II}$ well (\textit{second column}). E-valuators are excluded from this plot as they only allow to control the false alarm rate $\R^{I}$.}
    \label{fig:monitors_md}
\end{figure*}

\paragraph{Controlling Missed Detection Risk on Factuality}
We now select the threshold to give a guarantee on the missed detection risk (one minus
power) rather than on the false alarm risk. Keeping $\R^{II}$ low ensures a certain
power level, at the cost of admitting more false alarms. We carry out this calibration
in the mathematical reasoning setting of \Cref{sec:exp_factuality}. Notably, to our knowledge no existing LLM monitoring tool provides guarantees on the missed detection rate---including the e-valuator framework---meaning no direct baseline exists for this setting. \Cref{fig:monitors_md}
displays the results. The missed detection risk remains controlled at the prescribed
level. At low missed detection risk, the false alarm rate is correspondingly
high. For Claude---which produces a larger fraction of safe samples---the
false alarm rate is higher, while for Mistral---which produces more incorrect
samples---the false alarm rate is lower.

\section{The E-valuator framework}
\label{app:e_valuator}

We briefly summarize the \emph{E-valuator} framework from \citet{sadhuka2025valuator} below. Given a prompt, the LLM produces a variable-length trajectory $o_{1:T}$, and after each step $t$ a verifier is employed to return the score $s_t = p_{\psi}(y \mid x, o_{1:t})$ based on the (partial) trajectory thus far. The full score sequence $s_{1:T} = (s_1, \dots, s_T)$ is paired with a final label $y \in \{0,1\}$ indicating whether the final output is correct. In \citet{sadhuka2025valuator}, the monitoring task is then cast as a \emph{sequential hypothesis test} that distinguishes successful trajectories from unsuccessful ones via the class-conditional distributions of verifier scores:
\begin{equation*}
H_0: s_{1:T} \sim P_1 \qquad \text{vs.} \qquad H_1: s_{1:T} \sim P_0,
\end{equation*}
where $P_1$ has density $p_1(s_{1:T}) = p(s_{1:T} \mid y = 1)$, i.e.~correct, and $P_0$ has density $p_0(s_{1:T}) = p(s_{1:T} \mid y = 0)$, i.e.~incorrect. For a new LLM trajectory, the procedure evaluates a statistic $M_t$ online, and rejects the null once $M_t$ crosses a threshold $c_\alpha$. The target guarantee is \emph{anytime-valid} false-alarm control, or
\begin{equation*}
\mathbb{P}_{H_0} \bigl(\exists\, t \geq 1: M_t \ge c_\alpha \bigr) \le \alpha,
\end{equation*}
meaning that a true successful trajectory is incorrectly flagged at any time with probability at most $\alpha$, for any unknown sequence length $T$. 

\paragraph{E-valuator-anytime.} The key mechanism is an \emph{e-process} instantiated by the sequential density ratio $M_t = p_0(s_{1:t}) / p_1(s_{1:t})$, which forms a test martingale under $H_0$ and therefore admits the threshold $c_\alpha = 1/\alpha$ with strict time-uniform error control via Ville's inequality \cite{ramdas2025hypothesis}. Intuitively, $M_t$ grows when the observed score history looks more consistent with the distribution of unsuccessful trajectories than with that of successful ones. Since $p_0(s_{1:t})$ and $p_1(s_{1:t})$ are unknown in practice, \citet{sadhuka2025valuator} estimate their ratio discriminatively using Bayes' rule. Accordingly, for each step $t$ a probabilistic classifier $\hat f_t(s_{1:t}) \approx p(y = 1 \mid s_{1:t})$ is trained on calibration data, yielding a plug-in statistic that approximates the true likelihood ratio as
\begin{equation*}
\hat M_t = \frac{1-\hat f_t(s_{1:t})}{\hat f_t(s_{1:t})}
\frac{\hat \pi_1}{1 -\hat \pi_1},
\end{equation*}
where $\pi_1 \approx p(y = 1)$ forms a base rate. 

\paragraph{E-valuator-PAC.} To empirically improve power, \citet{sadhuka2025valuator} also suggest a data-driven relaxation of the rejection threshold $1/\alpha$, which can be conservative in finite-horizon deployments. Leveraging an additional held-out calibration split of successful trajectories, they estimate the null distribution of the running maximum $\max_t \hat M_t$, and subsequently set $c_\alpha$ to a high-probability upper bound on its $(1-\alpha)$-quantile. This yields a \emph{probably-approximately-correct}-type (PAC) threshold with guarantee
\begin{equation*}
\mathbb{P}_{D_{\text{cal}}} \left(\mathbb{P}_{H_0} \bigl( \exists\, t \geq 1: \hat M_t \ge c_\alpha \mid D_{\text{cal}} \bigr) \le \alpha \right) \ge 1-\delta,
\end{equation*}
which weakens the exact anytime-valid guarantee to a high-probability guarantee over the calibration sample. Unlike the exact threshold, false alarm then need not hold for every realized calibration draw, but the alarm trigger can be substantially less conservative and therefore more powerful in practice.

\end{document}